\documentclass[final,nonatbib]{article} 
\usepackage{nips_2018}

\usepackage{amsmath,amsfonts,bm}

\def\eqref#1{equation~\ref{#1}}

\def\1{\bm{1}}

\DeclareMathAlphabet{\mathsfit}{\encodingdefault}{\sfdefault}{m}{sl}
\SetMathAlphabet{\mathsfit}{bold}{\encodingdefault}{\sfdefault}{bx}{n}

\usepackage{hyperref}
\usepackage{url}
\usepackage{graphicx}
\graphicspath{ {figs/} }
\usepackage{tikz}
\usepackage{subcaption}
\usepackage{enumitem}
\usepackage{cprotect}

\title{Spectral Multigraph Networks for Discovering and Fusing Relationships in Molecules}

\author{Boris Knyazev\thanks{This work was done while the author was an intern at SRI International.}  \ $^{1,2}$ \& Graham W.~Taylor$^{1,2,3}$\\
	$^1$University of Guelph, Canada \\
	$^2$Vector Institute for Artificial Intelligence, Canada \\
	$^3$Canadian Institute for Advanced Research\\
	\texttt{\{bknyazev,gwtaylor\}@uoguelph.ca} \\
	\And
	Xiao Lin \& Mohamed R.~Amer \\
	SRI International \\
	201 Washington Rd\\
	Princeton, NJ08540, USA\\
	\texttt{\{xiao.lin,mohamed.amer\}@sri.com}
}

\begin{document}

	\maketitle
	\begin{abstract}
		Spectral Graph Convolutional Networks (GCNs) are a generalization of convolutional networks to learning on graph-structured data.
		Applications of spectral GCNs have been successful, but \textit{limited} to a few problems where the graph is fixed, such as shape correspondence and node classification. In this work, we address this limitation by revisiting a particular family of spectral graph networks, Chebyshev GCNs, showing its efficacy in solving graph classification tasks with a variable graph structure and size. Chebyshev GCNs restrict graphs to have at most one edge between any pair of nodes. To this end, we propose a novel \textit{multigraph} network that learns from multi-relational graphs. We model learned edges with abstract meaning and experiment with different ways to fuse the representations extracted from annotated and learned edges, achieving competitive results on a variety of chemical classification benchmarks.
	\end{abstract}
	\section{Introduction}\label{sec:intro}
	Convolutional Neural Networks (CNNs) have seen wide success in domains where data is restricted to a Euclidean space. These methods exploit properties such as stationarity of the data distributions, locality and a well-defined notation of translation, but cannot model data that is non-Euclidean in nature. Such structure is naturally present in many domains, such as chemistry, physics, social networks, transportation systems, and 3D geometry, and can be expressed by graphs~\cite{bronstein2017geometric, hamilton2017representation}. By defining an operation on graphs analogous to convolution, Graph Convolutional Networks (GCNs) have extended CNNs to graph-based data. The earliest methods performed convolution in the spectral domain~\cite{bruna2013spectral}, but subsequent work has proposed generalizations of convolution in the spatial domain. There have been multiple successful applications of GCNs to node classification~\cite{velickovic2017graph} and link prediction~\cite{schlichtkrull2018modeling}, whereas we target graph classification similarly to~\cite{simonovsky2017dynamic}.

	Our focus is on multigraphs, a graph that is permitted to have multiple edges. Multigraphs are important in many domains, such as chemistry and physics.
	The challenge of generalizing convolution to graphs and multigraphs is to have anisotropic convolution kernels (such as edge detectors). Anisotropic models, such as MoNet \cite{monti2017geometric} and SplineCNN~\cite{fey2018splinecnn}, rely on coordinate structure, work well for vision tasks, but are suboptimal for non-visual graph problems. Other general models exist~\cite{gilmer2017neural, battaglia2018relational}, but making them efficient for a variety of tasks conflicts with the ``no free lunch theorem''.

	Compared to non-spectral GCNs, spectral models have filters with more global support, which is important for capturing complex relationships. We rely on Chebyshev GCNs (ChebNet)~\cite{defferrard2016convolutional} that enjoy an explicit control of receptive
	field size. Even though it was originally derived from spectral methods~\cite{bruna2013spectral}, it does not suffer from their main shortcoming --- sensitivity of learned filters to graph size and structure.

	\textbf{Contributions:} We propose a scalable spectral GCN that learns from multigraphs by capturing multi-relational graph paths as well as multiplicative and additive interactions to reduce model complexity and learn richer representations. We also learn new abstract relationships between graph nodes, beyond the ones annotated in the datasets. To our knowledge, we are the first to demonstrate that spectral methods can efficiently solve problems with variable graph size and structure, where this kind of method is generally believed not to perform well.

	\section{Multigraph Convolution}\label{sec:multigraph}
	While we provide the background to understand our model, a review of spectral graph methods is beyond the scope of this paper. Section~\ref{sec:spectral_graph_conv_details} of the Appendix reviews spectral graph convolution.\looseness=-1
	\subsection{Approximate spectral graph convolution}
	\label{sec:spectral_graph_conv}
	We consider an undirected, possibly disconnected, graph $\cal G =(V, E)$ with $N$ nodes, $\cal V$, and edges, $\cal E$, having values in range $[0, 1]$. Nodes $v_i \in \cal V$ usually represent specific semantic concepts such as atoms in a chemical compound or users in a social network. Nodes can also denote abstract blocks of information with common properties, such as superpixels in images. Edges $e_{ij} \in \cal E$ define the relationships between nodes and the scope over which node effects may propagate.

	In spectral graph convolution~\cite{bruna2013spectral}, the filter $g \in R^N$ is defined on an entire input space.
	Although it makes filters global, which helps to capture complex relationships, it is also desirable to have local support since the data often have local structure and since we want to learn filters independent on the input size $N$ to make the model scalable.

	To address this issue, we can model this filter as a function of eigenvalues $\Lambda$ (which is assumed to be constant) of the normalized symmetric graph Laplacian $L$: $g = g(\Lambda)$. We can then approximate it as a sum of $K$ terms using the Chebyshev expansion, where each term $T_k(\Lambda) = 2 \Lambda T_{k-1}(\Lambda) - T_{k-2}(\Lambda)$ contains powers $\Lambda^k$. Finally, we apply the property of eigendecomposition:
	\begin{equation}
	\label{eq:eigen_property_power}
	L^k=(U \Lambda U^T)^k = U \Lambda^k U^T.
	\end{equation}
	By combining this property with the Chebyshev expansion of $g(\Lambda)$, we exclude eigenvectors $U \in \mathbb{R}^{N \times N}$, that are often infeasible to compute, from spectral graph convolution, and instead express the convolution as a function of graph Laplacian $L$.
	In general, for the input $X \in \mathbb{R}^{N \times X_{in}}$ with $N$ nodes and $X_{in}$-dimensional features in each node, the approximate convolution is defined as:
	\begin{equation}
	\label{eq:cheb_conv_general}
	Y = \bar{X} \Theta,
	\end{equation}
	where $\bar{X} \in \mathbb{R}^{N \times X_{in}K}$ are features projected onto the Chebyshev basis $T_k(\tilde{L})$ and concatenated for all orders $k \in [0, K-1]$ and $\Theta \in \mathbb{R}^{X_{in}K \times X_{out}}$ are trainable weights, where $\tilde{L} = L - I$.

	This approximation scheme was proposed in~\cite{defferrard2016convolutional}, and Eq.~\ref{eq:cheb_conv_general} defines the convolutional layer in the Chebyshev GCN (ChebNet), which is the basis for our method. Convolution is an essential computational block in graph networks, since it permits the gradual aggregation of information from neighboring nodes. By stacking the operator in~Eq.~\ref{eq:cheb_conv_general}, we capture increasingly larger neighborhoods and learn complex relationships in graph-structured data.
	\subsection{Graphs with variable structure and size}
	\label{sec:graph_structure_size}
	The approximate spectral graph convolution (Eq.~\ref{eq:cheb_conv_general}) enforces spatial locality of the filters by controlling the order of the Chebyshev polynomial $K$. Importantly, it reduces the computational complexity of spectral convolution from $O(N^2)$ to $O(K|\cal{E}|)$, making it much faster in practice assuming the graph is sparsely connected and sparse matrix multiplication is implemented.
	In this work, we observe an important byproduct of this scheme: that learned filters become less sensitive to changes in graph structure and size due to excluding the eigenvectors $U$ from spectral convolution, so that learned filters are not tied to $U$.

	\begin{figure}[]
		\begin{center}
			\vspace{-5pt}
			\begin{tabular}{ccc}
				\includegraphics[width=0.25\textwidth, trim={0.8cm 0cm 0cm 0cm}, clip]{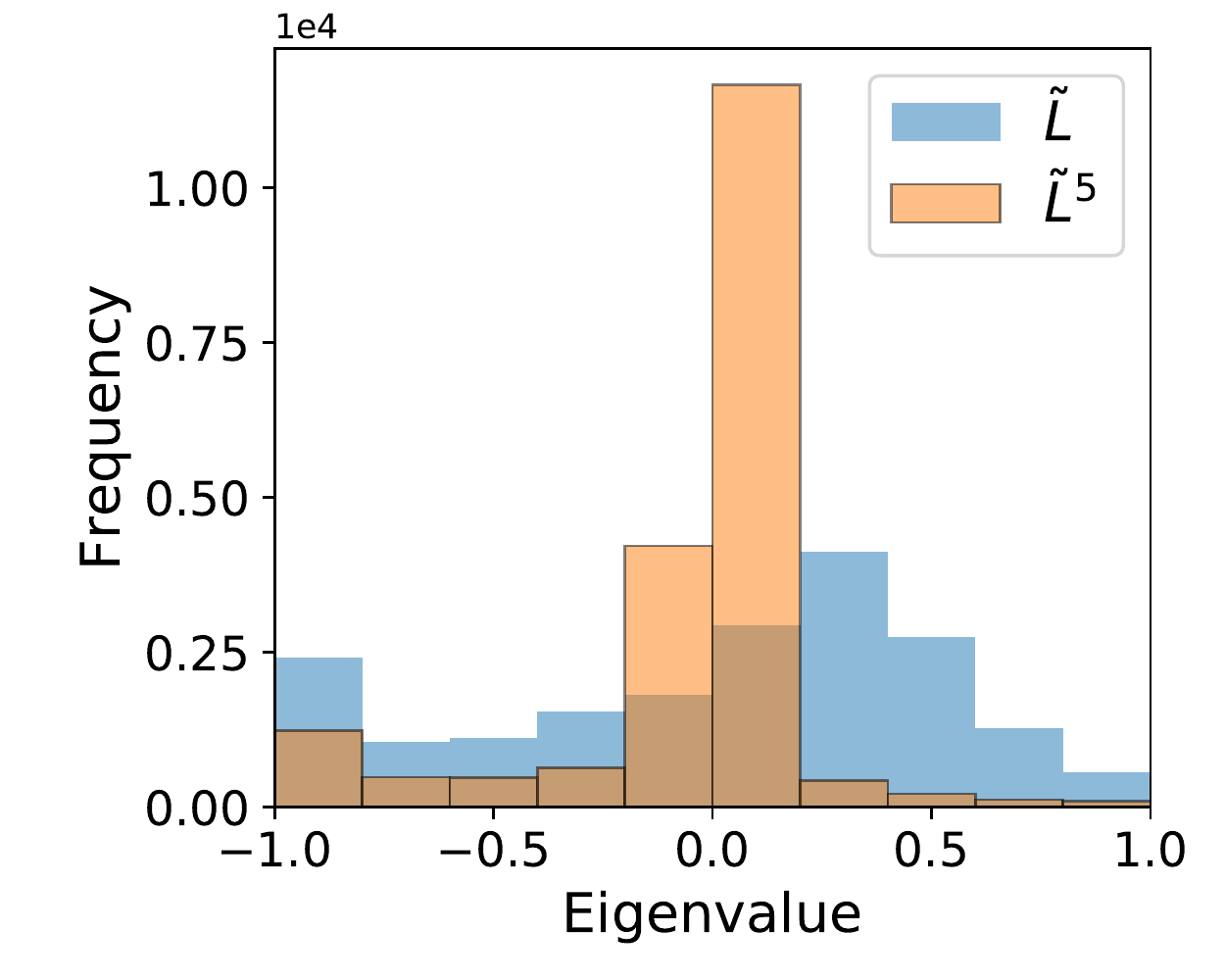} &
				\includegraphics[width=0.25\textwidth, trim={0.8cm 0cm 0cm 0cm}, clip]{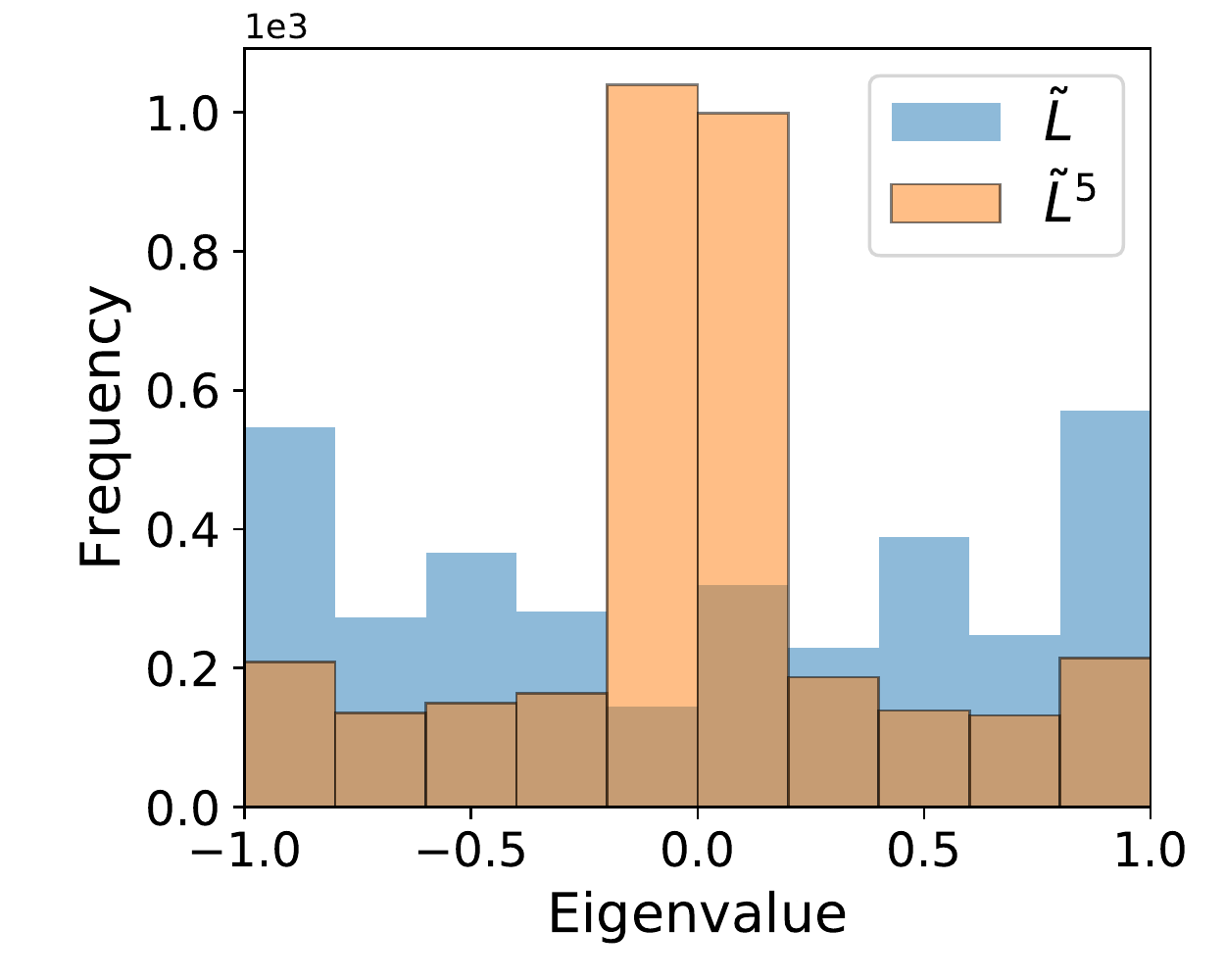} &
				\includegraphics[width=0.25\textwidth, trim={0.8cm 0cm 0cm 0cm}, clip]{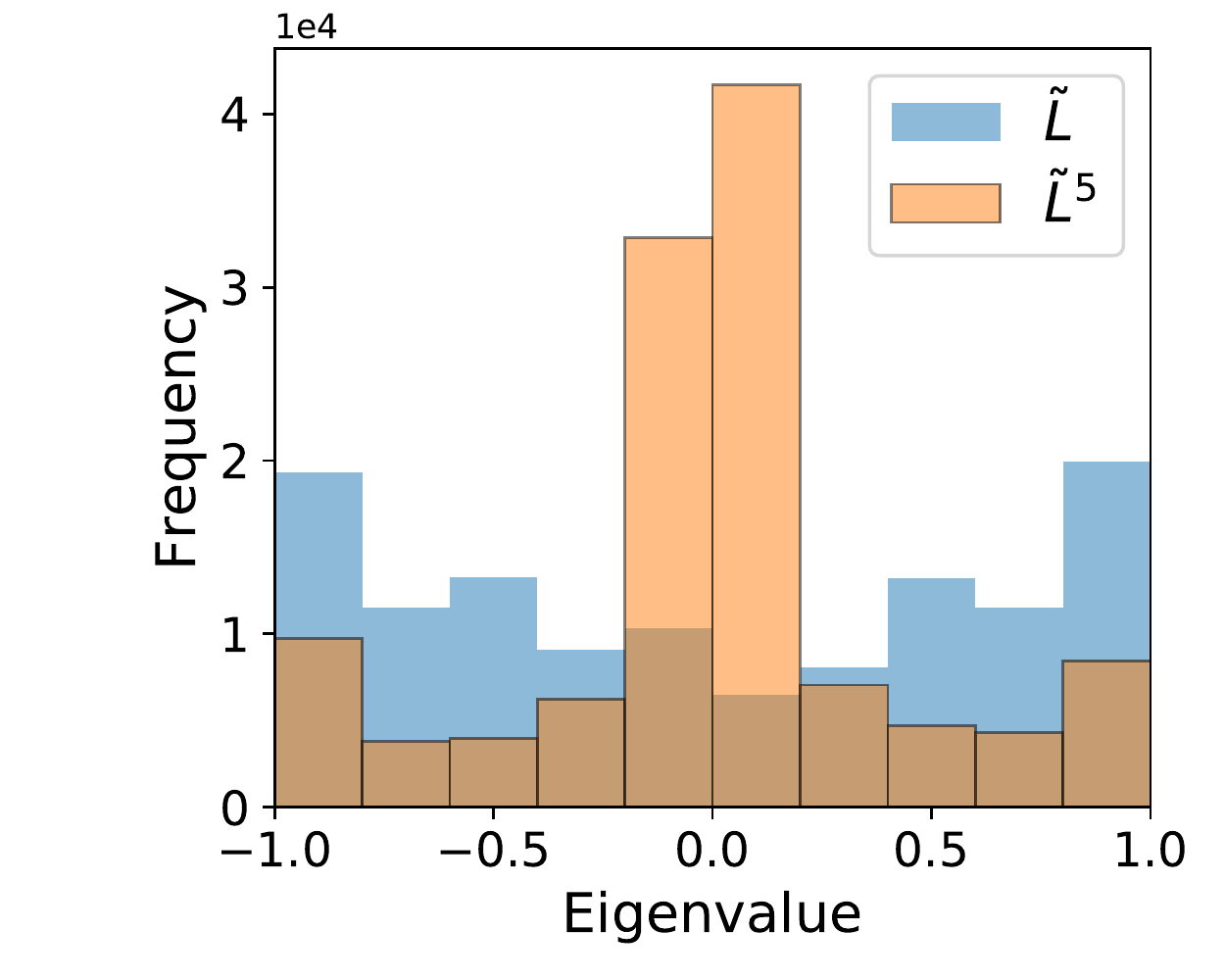} \\
				(a) ENZYMES & (b) MUTAG & (c) NCI1
				\vspace{-10pt}
			\end{tabular}
		\end{center}
		\caption{Histograms of eigenvalues of the rescaled graph Laplacian $\tilde{L}$ for the (a) ENZYMES, (b) MUTAG and (c) NCI1 datasets. Due to the property of eigendecomposition ($\tilde{L}^k=U \tilde{\Lambda}^k U^T$) the distribution of eigenvalues shrinks when we take powers of $\tilde{L}$ to compute the approximate spectral graph convolution (Eq.~\ref{eq:cheb_conv_general}).
		}
		\label{fig:eigen_distr}
	\end{figure}

	The only assumption that still makes a trainable filter $\hat{g}$ sensitive to graph structure is that we model it as a function of eigenvalues $\hat{g}(\Lambda)$. However, the distribution of eigenvalues of the normalized Laplacian is concentrated in a limited range, making it a weaker dependency on graphs than the spectral convolution via eigenvectors, so that learned filters generalize well to new graphs. Moreover, since we use powers of $\tilde{L}$ in performing convolution (Eq.~\ref{eq:cheb_conv_general}),
	the distribution of eigenvalues $\tilde{\Lambda}$ further contracts due to exponentiation of the middle term on the RHS of Eq.~\ref{eq:eigen_property_power}. We believe that this effect accounts for the robustness of learned filters to changes in graph size or structure (Figure~\ref{fig:eigen_distr}).

	\subsection{Graphs with multiple relation types}
	\label{sec:edge_fusion_methods}
	In the approximate spectral graph convolution (Eq.~\ref{eq:cheb_conv_general}), the graph Laplacian $\tilde{L}$ encodes a single relation type between nodes. Yet, a graph may describe many types of distinct relations.
	In this section, we address this limitation by extending Eq.~\ref{eq:cheb_conv_general} to a multigraph, i.e.~a graph with multiple ($R \geq 1$) edges (relations) between the same nodes encoded as a set of graph Laplacians $\{\tilde{L}^{(r)}\}_1^R$, where $R$ is an upper bound on the number of edges per dyad.
	Extensions to a multigraph can also be applied to early spectral models~\cite{bruna2013spectral} but, since ChebNet was shown to be superior in downstream tasks, we choose to focus on the latter model.

\begin{figure}[]
		\begin{center}
			\begin{tabular}{ccc}
				{\includegraphics[width=0.26\textwidth,trim={2cm 17.5cm 22cm 1.5cm}, clip]{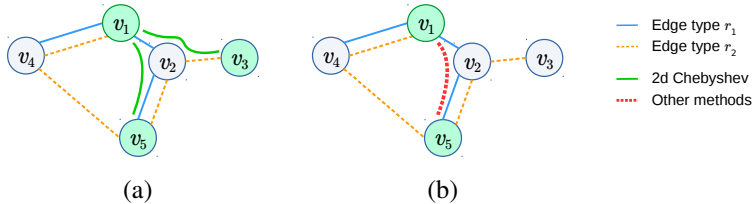}} &
				{\includegraphics[width=0.26\textwidth,trim={6.7cm 17.5cm 17.3cm 1.5cm}, clip]{edge_fusion.pdf}} &
				{\includegraphics[width=0.15\textwidth,trim={12cm 17cm 13cm 1.5cm}, clip]{edge_fusion.pdf}} \\
				(a) & (b) &
			\end{tabular}
		\end{center}
		\caption{Comparison of (a) the fusion method based on a two-dimensional (2d) Chebyshev polynomial (Eq.~\ref{eq:2d_cheb},~\ref{eq:2d_cheb_conv}) to (b) other proposed methods in case of a 2-hop filter (a filter averaging features of nodes located two edges away from the filter center - $v_1$ in this case). Note that (a) can leverage multi-relational paths and the filter centered at node $v_1$ can access features of the node $v_3$, which is not possible for other methods (b). In this work, edge type $r_1$ can denote annotated relations, while $r_2$ can denote learned ones (Eq.~\ref{eq:edge_predict}). We also allow for three and more relation types.}
		\label{fig:edge_fusion}
	\end{figure}

	\paragraph{Two dimensional Chebyshev polynomial.}
	The Chebyshev polynomial used in Eq.~\ref{eq:cheb_conv_general} (see Section~\ref{sec:spectral_graph_conv_details} in Appendix for detail) can be extended for two variables (relations in our case) similarly to bilinear models, e.g.~as in~\cite{omar2010two}:
	\begin{equation}
	\label{eq:2d_cheb}
	T_{ij}(\tilde{L}^{(r_1)},\tilde{L}^{(r_2)}) = T_i(\tilde{L}^{(r_1)})T_j(\tilde{L}^{(r_2)}), i,j = 0,...,K-1,
	\end{equation}

	and, analogously, for more variables. For $R=2$, the convolution is then defined as:
	\begin{equation}
	\label{eq:2d_cheb_conv}
	Y = [\bar{X}_{0,0}, \bar{X}_{0,1},...,\bar{X}_{i,j},..., \bar{X}_{K-1,K-1}] \Theta,
	\end{equation}

	where $\bar{X}_{i,j} = T_i(\tilde{L}^{(r_1)})T_j(\tilde{L}^{(r_2)}) X $.
	In this case, we allow the model to leverage graph paths consisting of multiple relation types (Figure~\ref{fig:edge_fusion}).
	This flexibility, however, comes at a great computational cost, which is prohibitive for a large number of relations $R$ or large order $K$ due to exponential growth of the number of parameters: $\Theta \in \mathbb{R}^{X_{in}K^R X_{out}}$. Moreover, as we demonstrate in our experiments, such multi-relational paths do not necessary lead to better performance.

	\paragraph{Multiplicative and additive fusion.}
	Motivated by multimodal fusion considered in the Visual Question Answering literature (e.g.~\cite{kim2016hadamard}), we propose the multiplicative operator:
	\begin{equation}
	\label{eq:multi_add_cheb_conv}
	Y = [f_{0}(\bar{X}^{(0)}) \odot f_{1}(\bar{X}^{(1)}) \odot ... \odot f_{R-1}(\bar{X}^{(R-1)})]  \Theta,
	\end{equation}

	where $f_r$ is a learnable differentiable transformation for relation type $r$ and $\bar{X}^{(r)}$ are features projected onto the Chebyshev basis $T_k(\tilde{L}^{(r)})$.
	In this case, node features interact in a multiplicative way. The advantage of this method is that it can learn separate $f_r$ for each relation and has fewer trainable parameters preventing overfitting, which is especially important for large $K$ and $R$.
	The element-wise multiplication $\odot$ in Eq.~\ref{eq:multi_add_cheb_conv} can be replaced with summation to perform additive fusion.

	 \paragraph{Shared projections.}
	 Another potential strength of the approach in Eq.~\ref{eq:multi_add_cheb_conv} is that we can further decrease model complexity by sharing parameters of $f_r$ between the relation types, so that the total number of trainable parameters does not depend on the number of relations $R$. Despite useful practical properties, as we demonstrate in the experiments, it is usually hard for a single shared $f_r$ to generalize between different relation types.

	\paragraph{Concatenating edge features.} A more straightforward approach is to concatenate features $\bar{X}^{(r)}$ for all $R$ relation types and learn a single matrix of weights $\Theta \in \mathbb{R}^{X_{in}KR \times X_{out}}$:
	\begin{equation}
	\label{eq:concat_cheb_conv}
	Y = [\bar{X}^{(0)}, \bar{X}^{(1)},...,\bar{X}^{(R-1)}] \Theta.
	\end{equation}
	This method, however, does not scale well for large $R$, since the dimensionality of $\Theta$ grows linearly with $R$. Note that even though multi-relational paths are not explicit in Eq.~\ref{eq:multi_add_cheb_conv} and \ref{eq:concat_cheb_conv}, for a \textit{multilayer} network, relation types will still communicate through node features. In Figure~\ref{fig:edge_fusion}, node $v_2$ will contain features of node $v_3$ after the first convolutional layer, so that in the second layer the filter centered at node $v_1$ will have access to features of node $v_3$ by accessing features of node $v_2$. Compared to the 2d polynomial convolution defined by Eq.~\ref{eq:2d_cheb_conv}, the concatenation-based, multiplicative and additive approaches require more layers to have a larger multi-relational receptive field.

	\section{Multigraph Convolutional Networks}
	\label{sec:model}
	A frequent assumption of current GCNs is that there is at most one edge between any pair of nodes in a graph. This restriction is usually implied by datasets with such structure, so that in
	many datasets, graphs are annotated with the single most important relation type, for example, whether two atoms in a molecule are bonded~\cite{wale2008comparison, duvenaud2015convolutional}. Meanwhile, data is often complex and nodes tend to have multiple relationships of different semantic, physical, or abstract meanings. Therefore, we argue that there could be other relationships captured by relaxing this restriction and allowing for multiple kinds of edges, beyond those annotated in the data.

	\subsection{Learning edges}
	\label{sec:learn_edges}
	Prior work (e.g.~\cite{schlichtkrull2018modeling,bordes2013translating}),
	proposed methods to learn from multiple edges, but similarly to the methods using a single edge type~\cite{kipf2016semi}, they leveraged only predefined (annotated) edges in the data.
	We devise a more flexible model, which, in addition to learning from an arbitrary number of predefined relations between nodes (see Section~\ref{sec:edge_fusion_methods}), learns abstract edges jointly with a GCN.
	We propose to learn a new edge $e^{(r)}_{ij}$ between any pair of nodes $v_i$ and $v_j$ with features $X_i$ and $X_j$ using a trainable similarity function:
	\begin{equation}
	\label{eq:edge_predict}
	e^{(r)}_{ij}=\frac{\exp\left(f_{\text{edge}}\left(X_i, X_j\right)\right)}{\sum_{k \in [1,|\cal V|]} \exp\left(f_{\text{edge}}\left(X_i, X_k\right)\right)},
	\end{equation}

	where the softmax is used to enforce sparse connections and $f_{\text{edge}}$ can be any differentiable function such as a multilayer perceptron in our work.
	This idea is similar to~\cite{henaff2015deep}, built on the early spectral convolution model~\cite{bruna2013spectral}, which learned an adjacency matrix, but targeted classification tasks for non graph-structured data (e.g.~document classification, with each document is represented as a feature vector). Moreover, we learn this matrix jointly with a more recent graph classification model~\cite{defferrard2016convolutional} and, additionally, efficiently fuse predefined and learned relations. Eq.~\ref{eq:edge_predict} is also similar to that of~\cite{velickovic2017graph}, which used this functional form to predict an attention coefficient $\alpha_{ij}$ for some \textit{existing} edge $e_{ij}$. The attention model can only strengthen or weaken some existing relations, but cannot form new relations. We present a more general model that makes it possible to connect previously disconnected nodes and form \textit{new} abstract relations.
	To enforce a symmetry of predicted edges we compute an average: $(e^{(r)}_{ij} + e^{(r)}_{ji}) / 2$.

	\subsection{Layer pooling versus global pooling}

	Inspired by convolutional networks, previous works~\cite{bruna2013spectral, defferrard2016convolutional,monti2017geometric,simonovsky2017dynamic,fey2018splinecnn} built an analogy of pooling layers in graphs, for example, using the Graclus clustering algorithm~\cite{dhillon2007weighted}.
	In CNNs, pooling is an effective way to reduce memory and computation, particularly for large inputs. It also provides additional robustness to local deformations and leads to faster growth of receptive fields. However, we can build a convolutional network without any pooling layers with similar performance in a downstream task~\cite{springenberg2014striving} --- it just will be relatively slow, since pooling is extremely cheap on regular grids, such as images.
	In graph classification tasks, the input dimensionality, which corresponds to the number of nodes $N=|\cal V|$, is often very small ($\sim10^2$) and the benefits of pooling are less clear. Graph pooling, such as in~\cite{dhillon2007weighted}, is also computationally intensive since we need to run the clustering algorithm for each training example independently, which limits the scale of problems we can address.
	Aiming to simplify the model while maintaining classification accuracy, we exclude pooling layers between conv.~layers and perform global maximum pooling (GMP) over nodes following the last conv.~layer. This fixes the size of the penultimate feature vector regardless of the number of nodes (Figure~\ref{fig:pipeline}).

	\section{Experiments}\label{sec:experiments}

	\subsection{Dataset details}
	We evaluate our model on five chemical graph classification datasets frequently used in previous work: NCI1 and NCI109~\cite{wale2008comparison}, MUTAG~\cite{debnath1991structure}, ENZYMES~\cite{schomburg2004brenda}, and PROTEINS~\cite{borgwardt2005protein}. For each dataset, there is a set of graphs with an arbitrary number of nodes $N=|\cal V|$ and undirected binary edges $|\cal E|$ of a single type ($R=1$) and each graph $\cal G$ has a single, categorical label that is to be predicted. Dataset statistics are presented in Table~\ref{table:networks} of the Appendix.

	Every graph represents some chemical compound labeled according to its functional properties.
	In NCI1, NCI109 and MUTAG, edges correspond to atom bonds (types of bonds) and vertices - to atom properties;	in ENZYMES, edges are formed based on spatial distance (edges connect nodes if those are neighbors along the amino acids sequence or if they are neighbors in space within the protein structure~\cite{borgwardt2005protein}); in PROTEINS, edges are similarly formed based on spatial distance between amino acids in proteins. Since edges in these datasets are not directly related to the features of nodes they connect, we expect that learned edges will enrich graph structure and improve graph classification.

	Node features $X$ are discrete in these datasets and represesented as one-hot vectors of length $X_{in}$. We do not use any additional node or edge attributes available for some of these datasets.

	These datasets vary in the number of graphs (188 - 4127), class labels (2~-~6) and the number of nodes in a graph (2~-~620) and, thereby, represent a comprehensive benchmark for our method. We follow the standard approach to evaluation~\cite{shervashidze2011weisfeiler,yanardag2015deep} and perform 10-fold cross-validation on these datasets. To minimize any random effects, we repeat experiments 10 times and report average classification accuracies together with standard deviations.

	\subsection{Architectural details and experimental setup}
	\begin{figure}[]
		\begin{center}
			\includegraphics[width=0.95\textwidth, trim={3cm 15cm 1.7cm 1.7cm}, clip]{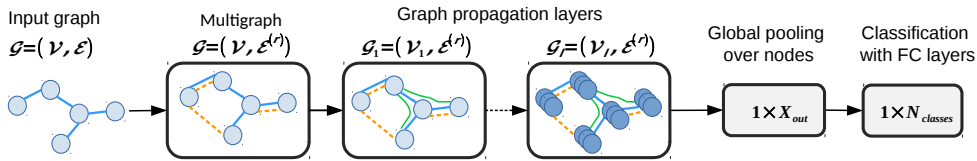}
		\end{center}
		\caption{Graph classification pipeline. Each $l\textsuperscript{th}$ convolutional layer in our model takes the graph ${\cal G}_l = ({\cal V}_l, {\cal E}^{(r)})$ and returns a graph with the same nodes and edges. Node features become increasingly global after each subsequent layer as the receptive field increases, while edges are propagated without changes. As a result, after several graph convolutional layers, each node in the graph contains information about its neighbors and the entire graph. By pooling over nodes we summarize the information collected by each node. Fully-connected layers follow global pooling to perform classification. Dashed orange edges denote connections learned as described in Section~\ref{sec:learn_edges}.
		}
		\label{fig:pipeline}
	\end{figure}

	In all experiments, we train a ChebNet with three graph convolutional layers followed by global max pooling (GMP) and 2 fully-connected layers (Figure~\ref{fig:pipeline}). Batch normalization (BN) and the ReLU activation are added after each layer, whereas dropout is added only before the fully-connected layers. Projections $f_r(\bar{X}^{(r)})$ in Eq.~\ref{eq:multi_add_cheb_conv} are modeled by a single layer neural network with $C=128$ hidden units and the $\tanh$ activation. The edge prediction function $f_{\text{edge}}$ (see Eq.~\ref{eq:edge_predict}, Section~\ref{sec:learn_edges}) is a two layer neural network with 128 hidden units (32 for PROTEINS), which acts on concatenated node features. Detailed network architectures are presented in Table~\ref{table:networks} of the Appendix.

	We train all models using the Adam optimizer~\cite{kingma2014adam} with learning rate of 0.001, weight decay of 0.0001, and batch size of 32, the learning rate is decayed after 25, 35, and 45 epochs and the models are trained for 50 epochs as in~\cite{simonovsky2017dynamic}.
	We run experiments for different fusion methods (Section~\ref{sec:edge_fusion_methods}) and Chebyshev orders $K$ in range from 2 to 6 (Section~\ref{sec:spectral_graph_conv}) and report the best results in Table~\ref{table:graph_class_results}.

	\newcommand\Tstrut{\rule{0pt}{2.6ex}}
	\newcommand\Bstrut{\rule[-0.9ex]{0pt}{0pt}}
	\newcommand{\std}[1]{{\scriptsize{$\pm$#1}}}
	\newcommand{\best}[1]{{\bfseries#1}}

	\begin{table}[t!]
		\caption{Chemical graph classification results (average accuracy and standard deviation in \%). Multigraph ChebNet obtains better results by leveraging two types of edges: annotated and learned, whereas all other models use only annotated edges. *We implemented MoNet, GCN and ChebNet. To make a fair comparison to Multigraph ChebNet, we use the same network architectures, batch-normalization, global max pooling. For MoNet, coordinates are defined using node degrees as in~\cite{monti2017geometric}. The top result across all methods for each dataset is bolded.}
		\label{table:graph_class_results}
		\small
		\begin{center}
			\begin{tabular}{lccccc}
				\textbf{Model}     & \textbf{NCI1} & \textbf{ NCI109} & \textbf{MUTAG} & \textbf{ENZYMES} & \textbf{PROTEINS}\Bstrut\\
				\hline
				WL~\cite{shervashidze2011weisfeiler} & 84.6\std{0.4} & 84.5\std{0.2} & 83.8\std{1.5} & 59.1\std{1.1} & $-$ \Tstrut \\
				WL-OA~\cite{kriege2016valid} & \best{86.1\std{0.2}} & \best{86.3\std{0.2}} & 84.5\std{1.7} & 59.9\std{1.1} & 76.4\std{0.4} \\
				structure2vec~\cite{dai2016discriminative} & 83.7 & 82.2 & 88.3 & 61.1 & $-$\\
				DGK~\cite{yanardag2015deep} & 80.3\std{0.5} & 80.3\std{0.3} & 87.4\std{2.7} & 53.4\std{0.9} & 75.7\std{0.5} \\
				PSCN~\cite{niepert2016learning} & 78.6\std{1.9} & $-$ & \best{92.6\std{4.2}} & $-$ & 75.9\std{2.8} \\
			 	ECC~\cite{simonovsky2017dynamic} & 83.8 & 82.1 & 88.3 & 53.5 & $-$\\
				DGCNN~\cite{zhang2018end} & 74.4\std{0.5} & $-$ & 85.8\std{1.7} & $-$ & 76.3\std{0.2} \\
				Graph U-Net~\cite{cangea2018towards} & $-$ & $-$ & $-$ & 64.2 & 75.5 \\
				DiffPool~\cite{ying2018hierarchical} & $-$ & $-$ & $-$ & 62.5 & 76.3 \\
				MoNet~\cite{monti2017geometric} - ours* & 69.8\std{0.2} & 70.0\std{0.3} & 84.2\std{1.2} & 36.4\std{1.2} & 71.9\std{1.2}\\
				GCN~\cite{kipf2016semi} - ours* & 75.8\std{0.7} & 73.4\std{0.4} & 76.5\std{1.4} & 40.7\std{1.8} & 74.3\std{0.5}\\
				\vspace{5pt}
				ChebNet~\cite{defferrard2016convolutional} - ours*& 83.1\std{0.4} & 82.1\std{0.2} & 84.4\std{1.6} & 58.0\std{1.4} & 75.5\std{0.4} \Bstrut\\
				\hline 
				Multigraph ChebNet & 83.4\std{0.4} & 82.0\std{0.3} & 89.1\std{1.4} & 61.7\std{1.3} & \best{76.5\std{0.4}} \Tstrut \\
			\end{tabular}

		\end{center}
	\end{table}

	\vspace{-3pt}
	\subsection{Results}
	\label{sec:results}

	Previous works typically show strong performance on one or two datasets out of five that we use (Table~\ref{table:graph_class_results}). In contrast, the Multigraph ChebNet, leveraging two relation types (annotated and learned, see Section~\ref{sec:learn_edges}), shows high accuracy across all datasets.
	On PROTEINS we outperform all previous methods, while on ENZYMES two recent works based on differentiable pooling~\cite{ying2018hierarchical, cangea2018towards} are better,
	however it is difficult to compare to their results without the standard deviation of accuracies. We	also obtain competitive accuracy on NCI1 outperforming DGK~\cite{yanardag2015deep}, PSCN~\cite{niepert2016learning}, and DGCNN~\cite{zhang2018end}.
	Importantly, the Multigraph ChebNet with two edge types, i.e.~predefined dataset annotations and the learned edges (Section~\ref{sec:learn_edges}) consistently outperforms the baseline ChebNet with a single edge, which shows efficacy of our approach and demonstrates the complementary nature of predefined and learned edges.
	Lower results on NCI1 and NCI109 can be explained by the fact that the node features in the graphs of these datasets are imbalanced with some features appearing only a few times in the dataset. This is undesirable for our method, which learns new edges based on features and the model can predict random values for unseen features.
	On MUTAG we surpass all but one method~\cite{niepert2016learning}. But in this case the dataset is tiny, consisting of 188 graphs and the margin from the top method is not statistically significant.

\begin{figure}[]
		\definecolor{col1}{HTML}{1F77B4}
		\definecolor{col2}{HTML}{FF7F0E}
		\definecolor{col3}{HTML}{2CA02C}
		\definecolor{col4}{HTML}{D62728}
		\definecolor{col5}{HTML}{9467BD}
		\definecolor{col6}{HTML}{8C564B}
		\definecolor{col7}{HTML}{E377C2}
		\begin{subfigure}{0.5\textwidth}
			\begin{flushleft}
				\includegraphics[width=\textwidth]{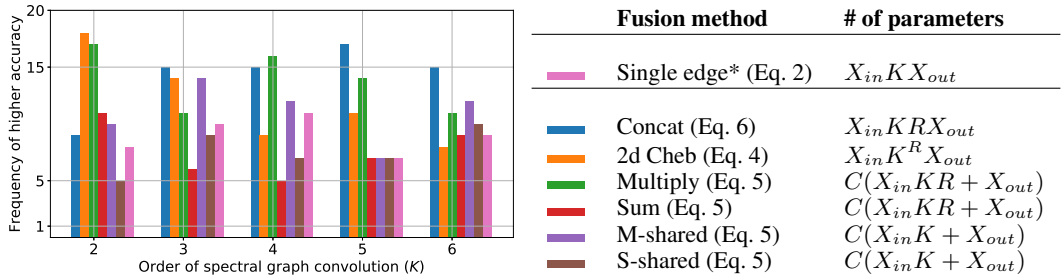}
			\end{flushleft}
		\end{subfigure}
		\begin{subfigure}{0.2\textwidth}
			\footnotesize
			\begin{tabular}{lll}
				& \textbf{Fusion method} & \textbf{\# of parameters} \\
				\hline \\
				\tikz \fill [col7] (0.7,0.3) rectangle (0.2,0.2); & Single edge* (Eq.~\ref{eq:cheb_conv_general}) & $X_{in}KX_{out}$ \\
				\hline \\
				\tikz \fill [col1] (0.7,0.3) rectangle (0.2,0.2); & Concat (Eq.~\ref{eq:concat_cheb_conv}) & $X_{in}KRX_{out}$  \\
				\tikz \fill [col2] (0.7,0.3) rectangle (0.2,0.2); & 2d Cheb (Eq.~\ref{eq:2d_cheb_conv}) 	& $X_{in}K^R X_{out}$ \\
				\tikz \fill [col3] (0.7,0.3) rectangle (0.2,0.2); & Multiply (Eq.~\ref{eq:multi_add_cheb_conv}) & $C(X_{in}KR + X_{out})$ \\
				\tikz \fill [col4] (0.7,0.3) rectangle (0.2,0.2); & Sum (Eq.~\ref{eq:multi_add_cheb_conv}) & $C(X_{in}KR + X_{out})$ \\
				\tikz \fill [col5] (0.7,0.3) rectangle (0.2,0.2); & M-shared (Eq.~\ref{eq:multi_add_cheb_conv}) & $C(X_{in}K + X_{out})$ \\
				\tikz \fill [col6] (0.7,0.3) rectangle (0.2,0.2); & S-shared (Eq.~\ref{eq:multi_add_cheb_conv}) & $C(X_{in}K + X_{out})$ \\
			\end{tabular}
		\end{subfigure}
		\cprotect\caption{Comparison of edge fusion methods for 10 folds.
			We observe that some methods perform well for lower order $K$, such as 2d Chebyshev convolution winning in 18/50 cases for $K=2$, while others perform better for higher $K$, such as \verb+Multiply-shared+. \verb+Multiply+ generally performs well across different $K$. All fusion methods, except for \verb+Sum+ and \verb+Sum-shared+, consistently outperform the \verb+Single-edge+ baseline.
			We also show the number of parameters in $\Theta$ for the Chebyshev convolution layer depending on the number of input features $X_{in}$, number of output features $X_{out}$, number of relation types $R$, order $K$ and some constant $C$ (number of hidden units in a projection layer $f_r$ in Eq.~\ref{eq:multi_add_cheb_conv}). *\verb+Single-edge+ denotes using only annotated edges. All other methods additionally use the second edge, learned based on node features.}%
		\label{fig:edge_fusion_compare}
	\end{figure}

	\paragraph{Evaluation of edge fusion methods.}
	We train a model using each of the edge fusion methods proposed in Section~\ref{sec:edge_fusion_methods} and report the summary of results in Figure~\ref{fig:edge_fusion_compare}. We count the number of times each method outperforms the others treating all 10 folds independently.
	As expected, graph convolution based on the two-dimensional Chebyshev polynomial is better for lower orders of $K$, since it exploits multi-relational graph paths, effectively increasing the receptive field of filters.
	However, for larger $K$, the model complexity becomes too high due to quadratic growth of the number of parameters and performance degrades.
	Sharing weights for multiplicative or additive fusion generally drops performance with a few exceptions in the multiplicative case. This implies that predefined and learned edges are of a different nature. It would be interesting to validate these fusion methods on a larger number of relation types.

\begin{figure}[]
		\begin{center}
			\includegraphics[width=0.5\textwidth]{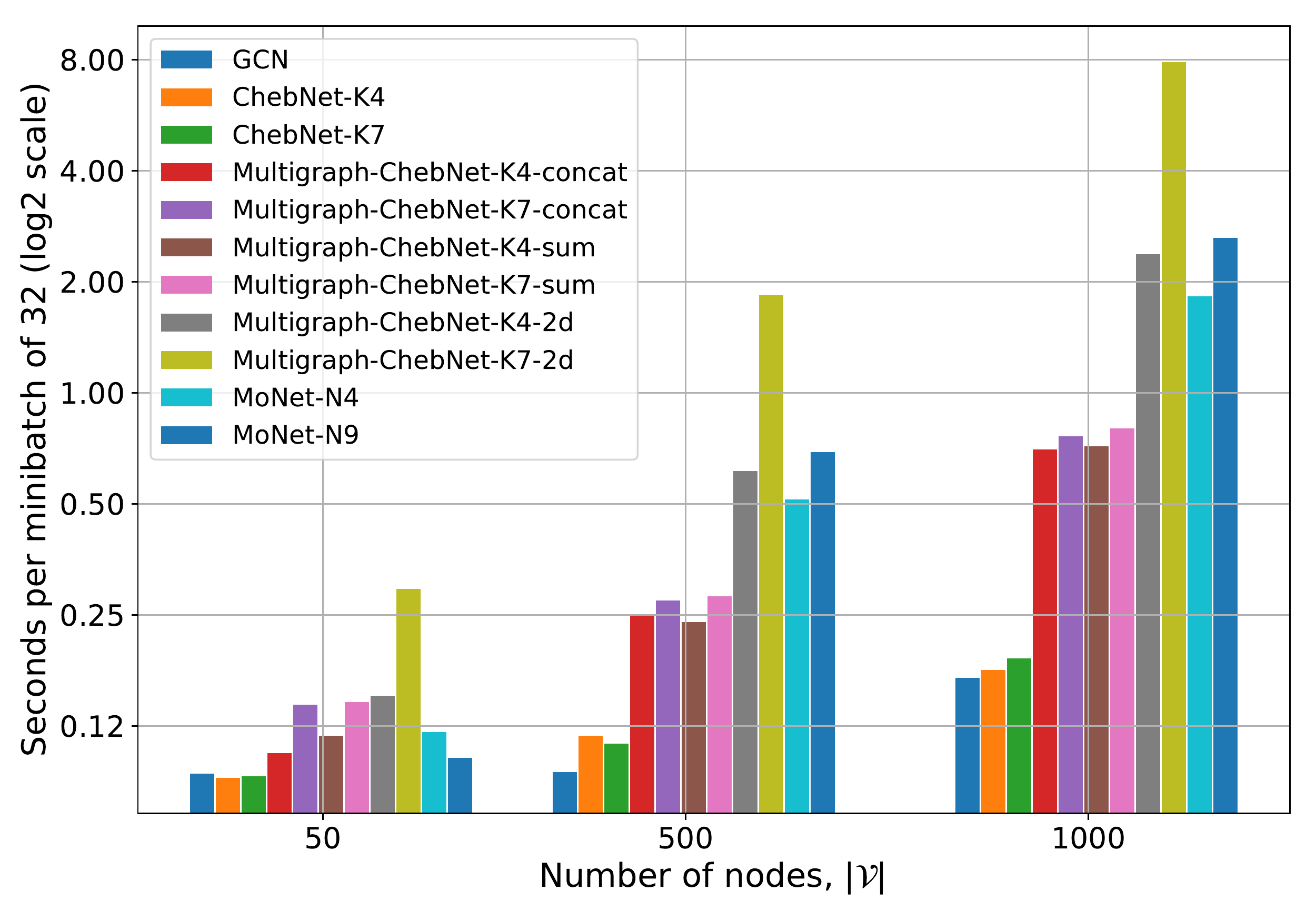}
		\end{center}
		\caption{Speed comparison of the baseline ChebNet, ChebNet with two edge types (Multigraph ChebNet), MoNet~\cite{monti2017geometric} and GCN~\cite{kipf2016semi}. MoNet-N$X$ refers to MoNet with $X$ filters.
		}
		\label{fig:speed}
	\end{figure}

	\paragraph{Speed comparison.}
	We compare forward pass speed of the proposed method to the baseline ChebNet, MoNet~\cite{monti2017geometric} and GCN~\cite{kipf2016semi} (Figure~\ref{fig:speed}). Analgously to~\cite{kipf2016semi}, we generate random graphs with $N$ nodes and $2N$ edges. ChebNet with 2 edge types (Multigraph ChebNet) is on average two times slower than the baseline ChebNet with a single annotated edge. MoNet is in turn two times slower than Multigraph ChebNet. Multigraph ChebNet with 2d edge fusion is the slowest due to exponential growth of parameters, while GCN is the fastest, although the gap with the baseline ChebNet is small. Therefore, we believe that Multigraph ChebNet with certain edge fusion methods provides a relatively fast, scalable and accurate model.

	\section{Related work and Discussion}
	\label{sec:works}

	Our method relies on a fast approximate spectral graph convolution known as ChebNet~\cite{defferrard2016convolutional}), which was designed for graph classification. A simplified and faster version of this model, Graph Convolutional Networks (GCN)~\cite{kipf2016semi}, which is practically equivalent to the ChebNet with order $K=1$, has shown impressive node classification performance on citation and knowledge graph datasets in the transductive learning setting. In all our experiments, we noticed that using more global filters (with larger $K$) is important (Tables~\ref{table:graph_class_results}).
	Other recent works~\cite{hamilton2017inductive, velickovic2017graph} also focus on node classification and, therefore, are not empirically compared to in this work.

	Recent work of~\cite{ying2018hierarchical} proposed a differentiable alternative to clustering-based graph pooling, showing strong results in graph classification tasks, but at the high computational cost. To alleviate this, a more scalable approach based on dropping nodes~\cite{graphunet2018, cangea2018towards} was introduced and can be integrated with our method to further improve results.

	Closely related to our work, \cite{monti2017geometric} formulated the generalized graph convolution model (MoNet) based on a trainable transformation to pseudo-coordinates, which led to learning anisotropic kernels and excellent results in visual tasks. However, in non-visual tasks, when coordinates are not naturally defined, the performance is worse (Table~\ref{table:graph_class_results}). Notably, the computational cost (both memory and speed) of MoNet is higher than for ChebNet due to the patch operator in~\cite[Eq. (9)-(11)]{monti2017geometric} (Figure~\ref{fig:speed}). The argument in favor of MoNet against ChebNet was the sensitivity of spectral convolution methods, including ChebNet, to changes in graph size and structure. We contradict this argument and show superior performance on chemical graph classification datasets.
	SplineCNN~\cite{fey2018splinecnn} is similar to MoNet and is good at classifying both graphs and nodes, but it is also based on pseudo coordinates and, therefore, potentially has the same shortcoming of MoNet. So, its performance on general graph classification problems where coordinates are not well defined is expected to be inferior.

	Another family of methods based on kernels~\cite{shervashidze2011weisfeiler, kriege2016valid} shows strong performance on chemical datasets,	but their application is limitted to small scale graph problems with discrete node features. Scalable extensions of	kernel methods to graphs with continuous features were
	proposed~\cite{niepert2016learning, yanardag2015deep}, but they showed weaker results.

	\section{Conclusion}
	\label{sec:conclusion}
	In this work, we address several limitations of current graph convolutional networks and show competitive graph classification results on a number of chemical datasets. First, we revisit the spectral graph convolution model based on the Chebyshev polynomial, commonly believed to inherit shortcomings of earlier spectral methods, and demonstrate its ability to learn from graphs of arbitrary size and structure. Second, we design and study edge fusion methods for multi-relational graphs, and show the importance of validating these methods for each task to achieve optimal performance. Third, we propose a way to learn new edges in a graph jointly with a graph classification model. Our results show that the learned edges are complimentary to edges already annotated, providing a significant gain in accuracy.

	\section*{Acknowledgments}
	This research was developed with funding from the Defense Advanced Research Projects Agency (DARPA). The views, opinions and/or findings expressed are those of the author and should not be interpreted as representing the official views or policies of the Department of Defense or the U.S. Government.
	The authors also acknowledge support from the Canadian Institute for Advanced Research and the Canada Foundation for Innovation.

	\bibliography{iclr2019_conference}
	\bibliographystyle{unsrt}
	\vfill

	\pagebreak
	\vfill

	\section*{Appendix}

	\subsection{Overview of spectral graph convolution and its approximation}
	\label{sec:spectral_graph_conv_details}

	Following the notation of~\cite{defferrard2016convolutional}, spectral convolution on a graph $\cal G$ having $N$ nodes is defined analogously to convolution in the Fourier domain (the convolution theorem) for some one-dimensional features over nodes $x \in \mathbb{R}^{N}$ and filter $g \in \mathbb{R}^N$ as~\cite{bruna2013spectral,bronstein2017geometric}:
	\begin{equation}
	\label{eq:spectral_conv_appdx}
	y = g \star x = U (U^Tg \odot U^Tx) = U \text{diag}(\hat{g}) U^Tx,
	\end{equation}
	where, $U$ are the eigenvectors of the normalized symmetric graph Laplacian, $L = I - D^{-1/2}AD^{-1/2}$, where $A$ is an adjacency matrix of the graph $\cal G$, $D$ are node degrees.
	$L=U \Lambda U^T$ follows from the definition of eigenvectors, where $\Lambda$ is a diagonal matrix of eigenvalues. The operator $\odot$ denotes the Hadamard product (element-wise multiplication), $\hat{g} = U^Tg$ and $\text{diag}(\hat{g})$ is a diagonal matrix with elements of $\hat{g}$ in the diagonal.

	The spectral convolution in (\ref{eq:spectral_conv_appdx}) can be approximated using the Chebyshev expansion, where $T_k(\Lambda) = 2 \Lambda T_{k-1}(\Lambda) - T_{k-2}(\Lambda)$ with $T_0(\Lambda) = 1$ and $T_1(\Lambda) = \Lambda$ (i.e. $T_k(\Lambda)$ terms contain powers $\Lambda^k$) and the property of eigendecomposition:
	\begin{equation}
	\label{eq:eigen_property_power_appdx}
	L^k=(U \Lambda U^T)^k = U \Lambda^k U^T.
	\end{equation}

	Assuming eigenvalues $\Lambda$ are fixed constants, filter $\hat{g}$ can be represented as a function of eigenvalues $\hat{g}(\Lambda)$, such that (\ref{eq:spectral_conv_appdx}) becomes:
	\begin{equation}
	\label{eq:spectral_conv_lambda_appdx}
	g \star x = U \hat{g}(\Lambda) U^T x.
	\end{equation}

	Filter $\hat{g}(\Lambda)$ can be then approximated as a Chebyshev polynomial of degree $K$ (a weighted sum of $T_k(\Lambda)$ terms).
	Substituting the approximated $\hat{g}(\Lambda)$ into Eq.~\ref{eq:spectral_conv_lambda_appdx} and exploiting Eq.~\ref{eq:eigen_property_power_appdx}, the approximate spectral convolution takes the form of (see in~\cite{defferrard2016convolutional, kipf2016semi} for further analysis and~\cite{hammond2011wavelets} for derivations):
	\begin{equation}
	\label{eq:cheb_conv_appdx}
	y = g \star x \approx U \left[ \sum^{K-1}_{k=0} \theta_k T_k(\tilde{\Lambda}) \right] U^T x = \sum^{K-1}_{k=0} \theta_k T_k(\tilde{L}) x = [\bar{x}_0,\bar{x}_1,...,\bar{x}_{K-1}] \theta,
	\end{equation}
	where $\tilde{L} = 2L/\lambda_{\max} - I$ is a rescaled graph Laplacian with $\lambda_{\max}$ as the largest eigenvalue of $L$, $\bar{x}_k=T_k(\tilde{L}) x \in \mathbb{R}^{N}$ are projections of input features onto the Chebyshev basis and $\theta=[\theta_0, \theta_1,...,\theta_{K-1}]$ are learnable weights shared across nodes.
	In this work, we further simplify the computation and fix $\lambda_{\max} = 2$ ($\lambda_{\max}$ varies from graph to graph), so that $\tilde{L} = L - I = - D^{-1/2}AD^{-1/2}$ and assume no loops in a graph. $\tilde{L}$ has the same eigenvectors $U$ as $L$, but its eigenvalues are $\tilde{\Lambda} = \Lambda - 1$.

	\subsection{Dataset statistics and network architectures}

	\begin{table}[htp]
		\caption{Dataset statistics and graph network architectures.
		These statistics can also be found in~\cite{KKMMN2016} along with the datasets themselves. $N=|{\cal V}|$ - number of nodes in a graph, $X_{in}$ - input dimensionality. GC - graph convolution layer, FC - fully connected layer, D - dropout.}
		\label{table:networks}
		\small
		\centering
		\setlength{\tabcolsep}{5pt}
		\begin{tabular}{lllllll}
			\textbf{Dataset}  & \textbf{\# graphs} & $\boldmath{|{\cal V}|_{\text{min}}}$ & $\boldmath{|{\cal V}|_{\text{max}}}$ & $\boldmath{|{\cal V}|_{\text{avg}}}$ & $\boldmath{X_{in}}$ & \textbf{Architecture} \\
			\hline \\
			NCI1 & 4110 & 3 & 111 & 29.87 & 37 & GC32-GC64-GC128-D0.1-FC256-D0.1-FC2 \\
			NCI109 & 4127 & 4 & 111 & 29.68 & 38 & GC32-GC64-GC128-D0.1-FC256-D0.1-FC2 \\
			MUTAG & 188 & 10 & 28 & 17.93 & 7 & GC32-GC32-GC32-D0.1-FC96-D0.1-FC2 \\
			ENZYMES & 600 & 2 & 126 & 32.63 & 3 & GC32-GC64-GC512-D0.1-FC256-D0.1-FC6 \\
			PROTEINS & 1113 & 4 & 620 & 39.06 & 3 & GC32-GC32-GC32-D0.1-FC96-D0.1-FC2 \\
		\end{tabular}
	\end{table}

	\vfill

\end{document}